# Evidence-invariant Sensitivity Bounds

**Silja Renooij** and **Linda C. van der Gaag**
Institute of Information and Computing Sciences, Utrecht University
P.O. Box 80.089, 3508 TB Utrecht, The Netherlands
{silja,linda}@cs.uu.nl

## Abstract

The sensitivities revealed by a sensitivity analysis of a probabilistic network typically depend on the entered evidence. For a real-life network therefore, the analysis is performed a number of times, with different evidence. Although efficient algorithms for sensitivity analysis exist, a complete analysis is often infeasible because of the large range of possible combinations of observations. In this paper we present a method for studying sensitivities that are invariant to the evidence entered. Our method builds upon the idea of establishing bounds between which a parameter can be varied without ever inducing a change in the most likely value of a variable of interest.

## 1 INTRODUCTION

The numerical parameters for a probabilistic network are generally estimated from statistical data or assessed by human experts in the domain of application. As a consequence of incompleteness of data and partial knowledge of the domain, the assessments obtained inevitably are inaccurate. Since the outcome of a probabilistic network is built from these assessments, it may be sensitive to the inaccuracies involved and, as a result, may even be unreliable.

The reliability of the outcome of a probabilistic network can be evaluated by subjecting the network to a sensitivity analysis. Such an analysis amounts to varying the assessments for one or more of its numerical parameters and investigating the effects on a probability of interest. Efficient algorithms are available that build upon the observation that the sensitivity of a probability of interest to parameter variation complies with a simple mathematical function; this sensitivity function basically expresses the probability of interest in terms of the parameter under study [4, 5]. Computing the constants for a sensitivity function requires just a limited number of network propagations.

The sensitivities revealed by a sensitivity analysis of a prior probabilistic network typically differ from those revealed by an analysis of the network after evidence has been entered. A complete sensitivity analysis of a real-life network would therefore involve performing multiple analyses, for different profiles of evidence. Such a complete analysis generally is infeasible, however, as a consequence of the many different possible profiles. Consider, as an example, a probabilistic network having 25 observable variables with 3 values each. For this network, there are some $10^{14}$ different combinations of observations, or evidence profiles. A complete sensitivity analysis of the network would require a number of network propagations that is at least of the same order of magnitude.

The above example serves to demonstrate the need for methods that provide insight into sensitivities and their consequences, without actually performing sensitivity analyses for the full range of evidence profiles. Recent results show that the change in a probability of interest, that is occasioned by a shift in a given parameter, can be bounded without knowledge of the network under study [2, 3]. The provided bounds depend, more specifically, just on the original values of the parameter and of the probability of interest. In this paper, we argue that these bounds actually are bounds on a sensitivity function and are built from sensitivity functions themselves. Based upon this observation, we provide an upper bound on the effect of small shifts in a parameter, that is, on its sensitivity value. We further establish lower bounds on the deviation that is allowed for a parameter from its original value, before the most likely value of a variable of interest may change. These bounds can again be established without knowledge of the network under study. The bounds moreover are evidence-invariant in the sense that they hold for large ranges of profiles.

The paper is organised as follows. In Section 2, we present some preliminaries concerning sensitivity functions. In Section 3, we introduce bounds on sensitivity values; bounds on admissible deviations are discussed in Section 4. The paper ends with our conclusions and directions for further research in Section 5.

## 2 SENSITIVITY FUNCTIONS

Sensitivity analysis of a probabilistic network amounts to establishing, for each of the network's numerical parameters, the *sensitivity function* that expresses the probability of interest in terms of the parameter under study. In the sequel, we denote the probability of interest by $\Pr(A = a \mid e)$, or $\Pr(a \mid e)$ for short, where $a$ is a specific value of the variable $A$ of interest and $e$ denotes the available evidence. The network's parameters are denoted by $x = p(b_i \mid \pi)$, where $b_i$ is a value of a variable $B$ and $\pi$ is a combination of values for the parents of $B$. We use $f_{\Pr(a|e)}(x)$ to denote the sensitivity function that expresses the probability $\Pr(a \mid e)$ in terms of the parameter $x$; we often omit the subscript for the function symbol $f$, as long as ambiguity cannot occur.

Upon varying a single parameter $x = p(b_i \mid \pi)$, the other parameters $p(b_j \mid \pi)$, $j \neq i$, specified for the variable $B$ need to be *co-varied* to ensure that the parameters pertaining to the same conditional distribution still sum to one; each parameter $p(b_j \mid \pi)$ can in fact be seen as a function of the parameter $x$ under study. In the sequel, we assume that the parameters $p(b_j \mid \pi)$ are co-varied with $p(b_i \mid \pi)$ in such a way that their mutual proportional relationship is kept constant, that is,

$$p(b_j \mid \pi) \leftarrow p(b_j \mid \pi) \cdot \frac{1 - x}{1 - p(b_i \mid \pi)}$$

for $p(b_i \mid \pi) < 1$. This *proportional co-variation* has been shown to result in the smallest distance between the original and the new probability distribution upon variation [3].

Under the (standard) assumption of proportional co-variation, any sensitivity function $f_{\Pr(a|e)}(x)$ is a quotient of two functions that are linear in the parameter $x$ under study [1, 4]. The numerator of the quotient describes the probability $\Pr(a, e)$ as a function of the parameter $x$ and the denominator describes $\Pr(e)$ as a function of $x$. More formally, the function takes the form

$$f(x) = \frac{c_1 \cdot x + c_2}{c_3 \cdot x + c_4}$$

where the constants $c_j$, $j = 1, \ldots, 4$, are built from the assessments for the numerical parameters that are not being varied. Any sensitivity function is thus characterised by at most three constants. These constants can be feasibly determined from the network, for example by computing the probability of interest for up to three values for the parameter under study and solving the resulting system of equations [4], or by means of an algorithm that is closely related to junction-tree propagation [5].

The sensitivity function $f(x)$ provides for establishing the change in the probability of interest that is occasioned by a specific shift in the parameter $x$ under study. The effect of a small shift in the parameter is captured by the value $f'(x_0)$ of the first derivative of the function at the original value $x_0$ of the parameter; the absolute value of $f'(x_0)$ is called the *sensitivity value* of the parameter $x$ for the current probability of interest. Even if the probabilities of the various values of the variable of interest are not very sensitive to variation of the parameter $x$, a small shift in $x$ may still change the most likely value of this variable. For capturing the extent of the variation that can be applied to a parameter without changing the most likely value of the variable of interest, the concept of *admissible deviation* was introduced [6]. An admissible deviation for a variable of interest and a given parameter, is a pair of values $(\alpha, \beta)$ that describe the shifts to smaller values and to larger values, respectively, that are allowed in the parameter without inducing a change in the most likely value of the variable of interest. For a parameter with an original value of $x_0$, the admissible deviation $(\alpha, \beta)$ thus indicates that the parameter can be safely varied within the interval $[x_0 - \alpha, x_0 + \beta]$.

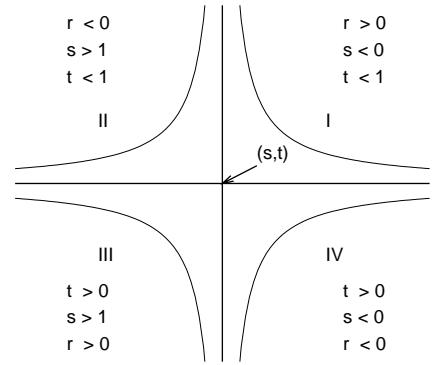

Figure 1: Hyperbolas and their constants (the constraints on $s$ and $t$ are specific for sensitivity functions).

A sensitivity function is either a *linear* function or a fragment of a *rectangular hyperbola*. A rectangular hyperbola takes the general form

$$f(x) = \frac{r}{x - s} + t$$

where, for a hyperbolic sensitivity function, we have that

$$s = -\frac{c_4}{c_3}, \quad t = \frac{c_1}{c_3}, \quad \text{and} \quad r = \frac{c_2 \cdot c_3 - c_1 \cdot c_4}{c_3^2}$$

The hyperbola has two branches and two asymptotes. Figure 1 illustrates the locations of the possible hyperbola branches relative to the two asymptotes. For $r < 0$, the branches lie in the second (II) and fourth (IV) quadrants relative to the asymptotes $x = s$ and $f(x) = t$; for $r > 0$, the branches are found in the first (I) and third (III) quadrants. Since any sensitivity function is continuous and well-defined for $x \in [0, 1]$, a hyperbolic sensitivity function is actually a fragment of one of the four possible hyperbola branches. As for each sensitivity function we further have that $0 \leq x \leq 1$ and $0 \leq f(x) \leq 1$, we observe that the vertical asymptote $x = s$ lies either to the left of $x = 0$ or to the right of $x = 1$. For any sensitivity function, therefore, we have that $s < 0$ or $s > 1$. In addition, we observe that the horizontal asymptote $f(x) = t$ lies below $f(1)$ for a first-quadrant function and below $f(0)$ for a second-quadrant function; we then have that $t < 1$. Similarly, the horizontal asymptote lies above $f(0)$ for a third-quadrant function and above $f(1)$ for a fourth-quadrant function, implying that $t > 0$. Note, for example, that for a type I function, negative values of $t$ are possible, and that for a type IV function, values of $t$ larger than 1 are possible.

## 3 BOUNDING SENSITIVITY VALUES

A sensitivity analysis is generally performed to investigate the extent to which a probability of interest $\Pr(a \mid e)$ can change as a consequence of a shift in the parameter $x$ from its original value $x_0$ to another value $x_1$. From the sensitivity function, the change in $\Pr(a \mid e)$ occasioned by the shift can be computed exactly. As the distance between $x_0$ and $x_1$ approaches zero, the change in $\Pr(a \mid e)$ is captured by the value $f'(x_0)$ of the first derivative of the sensitivity function, which can also be established exactly. In this section we show for both hyperbolic and linear sensitivity functions, that bounds on the change in $\Pr(a \mid e)$ and in the sensitivity value for $\Pr(a \mid e)$ can be established without actually knowing the sensitivity function. These bounds constitute the basis for analysing the effects of parameter variation for large ranges of evidence profiles.

### 3.1 HYPERBOLIC FUNCTIONS

We consider a parameter $x$ within a given probabilistic network. Suppose that $x_0$ is the value specified in the network for $x$, and that $p_0$ is the corresponding value of the probability of interest, which may be any prior or posterior probability. In the sequel, we refer to these values as the *original* values of the parameter and of the probability of interest, respectively. Without loss of generality, we assume that neither $x_0$ nor $p_0$ is equal to zero or one.

In their work on sensitivity analysis, Chan and Darwiche [3] established bounds on the new value $p_1$ of the probability of interest that results from varying the parameter $x$ from $x_0$ to $x_1$. Under the assumption of proportional co-variation, their bounds are given by

$$\frac{p_0 \cdot e^{-\delta}}{p_0 \cdot (e^{-\delta} - 1) + 1} \leq p_1 \leq \frac{p_0 \cdot e^{\delta}}{p_0 \cdot (e^{\delta} - 1) + 1}$$

where

$$\delta = \left| \ln \frac{x_1}{1 - x_1} - \ln \frac{x_0}{1 - x_0} \right|$$

While these bounds were stated for a fixed $x_1$, we observe that they are easily rephrased as bounds on the sensitivity function for all possible values of $x$, by replacing $x_1$ with $x$ and $p_1$ with $f(x)$. Note that the above bounds depend on the original values of the parameter and of the probability of interest, but are independent of any other aspect of the network under study. The bounds therefore apply to any pair $(x_0, p_0)$ for any network. Their computation, moreover, does not require any network propagations, except for establishing the value $p_0$ that corresponds with $x_0$.

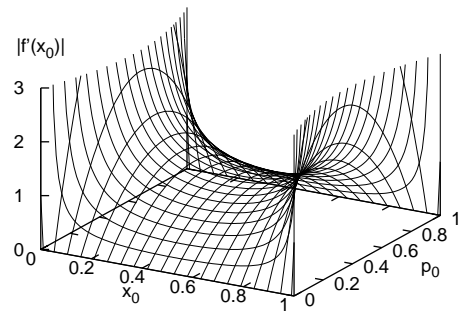

Figure 2: The upper bound on the sensitivity value of a hyperbolic sensitivity function, as a function of $x_0$ and $p_0$.

Chan and Darwiche [2] further established an upper bound on the sensitivity value of a parameter for a *binary* variable:

$$|f'(x_0)| \leq \frac{p_0 \cdot (1 - p_0)}{x_0 \cdot (1 - x_0)}$$

For ease of reference, the upper bound on the sensitivity value as a function of $x_0$ and $p_0$ is replicated in Figure 2. The figure reveals that large sensitivities are expected only for the more extreme values of $x_0$.

We recall that, for real-life probabilistic networks, often different profiles of evidence are studied in a sensitivity analysis. We demonstrate the possible uses of the above bound in view of such a thorough analysis. We consider, as an example, a parameter with an original value of $0.5$. We conclude from the above bound that any probability of interest will be quite insensitive to small shifts in this parameter, regardless of the evidence profile under study. Now suppose that we are interested in the sensitivity value for some probability of interest with an original value of $0.9$. For any profile that we can identify, using domain knowledge, as one that induces an increase of this probability, we observe that no parameter can upon variation induce a major change. On the other hand, for profiles that serve to decrease the original probability of interest, a shift in a parameter with a relatively extreme original value may induce a considerable change in the probability of interest given the profile. For such combinations of profiles and parameters, therefore, a more detailed analysis is required.

An in-depth study of the bounds on a sensitivity function established by Chan and Darwiche, reveals that these bounds are the maximum and minimum, respectively, of two rectangular hyperbolas.

**Proposition 3.1** *Let $x_0$ be the original value of a parameter $x$ and let $p_0$ be the corresponding probability of interest as before. Furthermore, let*

$$i(x) = \frac{p_0 \cdot (1 - x_0) \cdot x}{(p_0 - x_0) \cdot x + (1 - p_0) \cdot x_0}$$

*and*

$$d(x) = \frac{p_0 \cdot x_0 \cdot (1 - x)}{(1 - p_0 - x_0) \cdot x + p_0 \cdot x_0}$$

*Then, for any hyperbolic sensitivity function $f(x)$ with $f(x_0) = p_0$, we have that*

$$\min\{i(x_j), d(x_j)\} \leq f(x_j) \leq \max\{i(x_j), d(x_j)\}$$

*for all $x_j \in [0, 1]$.*

**Proof:** Any increasing hyperbolic sensitivity function $f(x)$ with $f(x_0) = p_0$ is bounded by an increasing hyperbola $i(x) = \frac{r_i}{x - s_i} + t_i$ with $i(0) = 0$, $i(1) = 1$, and $i(x_0) = p_0$, where $i(x)$ is a lower bound on $f(x)$ for $x < x_0$ and an upper bound on $f(x)$ for $x > x_0$. Any decreasing hyperbolic sensitivity function is bounded by a decreasing hyperbola $d(x) = \frac{r_d}{x - s_d} + t_d$ with $d(0) = 1$, $d(1) = 0$, and $d(x_0) = p_0$, where $d(x)$ is an upper bound on $f(x)$ for $x < x_0$ and a lower bound on $f(x)$ for $x > x_0$. From the three constraints per function, the constants of $i(x)$ and $d(x)$ can be computed:

$$s_i = \frac{x_0 - p_0 \cdot x_0}{x_0 - p_0}, \ \ t_i = 1 - s_i, \ \ r_i = s_i \cdot (1 - s_i)$$

and

$$s_d = \frac{p_0 \cdot x_0}{x_0 + p_0 - 1}, \ \ t_d = s_d, \ \ r_d = s_d \cdot (s_d - 1)$$

The result now follows immediately. □

From Proposition 3.1, we have that for $x \leq x_0$ the increasing hyperbola $i(x)$ is a lower bound on any sensitivity function $f(x)$ with $f(x_0) = p_0$ and the decreasing hyperbola $d(x)$ is an upper bound on $f(x)$; for the larger values of $x$, the two hyperbolas switch roles. It is now straightforward to show that the bounds from the proposition are equivalent to the bounds provided by Chan and Darwiche. By taking the first derivatives of the functions $i(x)$ and $d(x)$, moreover, we find the same upper bound on the sensitivity value as established by Chan and Darwiche. Their result therefore holds not just for the parameters of binary variables, but for the parameters of any variable in general.

## 3.2 LINEAR FUNCTIONS

In the previous section, we established hyperbolic bounds on a sensitivity function. Considering once more the general form of a sensitivity function, we note that it reduces to a linear function for $c_3 = 0$. If we know that a sensitivity function is linear in the parameter under study, we can establish tighter bounds than the hyperbolic bounds that we found before. We consider to this end an increasing linear function through $(0, 0)$ and $(1, 1)$. Note that this function passes through $(x_0, p_0)$ only if $x_0 = p_0$. For a linear function $i_l(x)$ to serve as a bound on a linear sensitivity function $f(x)$ with $f(x_0) = p_0$, it must either have $i_l(0) = 0$ or $i_l(1) = 1$. More specifically, for $x_0 > p_0$, the function has $i_l(0) = 0$; for $x_0 < p_0$, it has $i_l(1) = 1$. The function $i_l(x)$ thus has two points in common with an increasing hyperbolic bound. For either $x = 0$ or $x = 1$, however, the function's value will lie within the $\langle 0, 1 \rangle$-range. Similar observations pertain to a decreasing linear function. Figure 3 now illustrates that, although linear sensitivity functions are also bounded by the hyperbolas $i(x)$ and $d(x)$ from Proposition 3.1, tighter bounds can in fact be established.

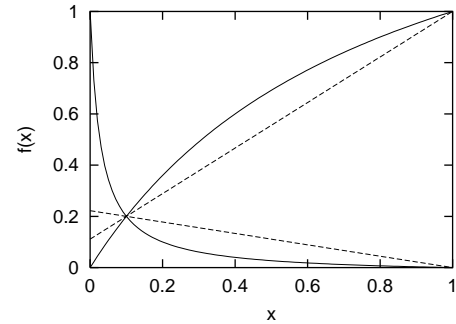

Figure 3: An example of linear and hyperbolic bounds on sensitivity functions through $(x_0, p_0) = (0.10, 0.20)$.

**Proposition 3.2** *Let $x_0$ be the original value of a parameter $x$ and let $p_0$ be the corresponding probability of interest as before. Furthermore, let*

$$i_l(x) = \begin{cases} \dfrac{p_0}{x_0} \cdot x, & \text{if } x_0 \geq p_0 \\ \dfrac{1 - p_0}{1 - x_0} \cdot x + \dfrac{p_0 - x_0}{1 - x_0}, & \text{otherwise} \end{cases}$$

*and*

$$d_l(x) = \begin{cases} \dfrac{p_0 - 1}{x_0} \cdot x + 1, & \text{if } x_0 \geq 1 - p_0 \\ \dfrac{-p_0}{1 - x_0} \cdot x + \dfrac{p_0}{1 - x_0}, & \text{otherwise} \end{cases}$$

*Then, for any linear sensitivity function $f(x)$ with $f(x_0) = p_0$, we have that*

$$\min\{i_l(x_j), d_l(x_j)\} \leq f(x_j) \leq \max\{i_l(x_j), d_l(x_j)\}$$

*for all $x_j \in [0, 1]$.*

**Proof:** Any increasing linear sensitivity function $f(x)$ with $f(x_0) = p_0$ is bounded by an increasing linear function $i_l(x)$ with $i_l(0) = 0$ for $x_0 \geq p_0$ or $i_l(1) = 1$ for $x_0 < p_0$, and $i_l(x_0) = p_0$, where $i_l(x)$ is a lower bound on $f(x)$ for $x < x_0$ and an upper bound on $f(x)$ for $x > x_0$. Any decreasing linear sensitivity function is bounded by a decreasing linear function $d_l(x)$ with $d_l(0) = 1$ for $x_0 \geq 1 - p_0$ or $d_l(1) = 0$ for $x_0 < 1 - p_0$, and $d_l(x_0) = p_0$, where $d_l(x)$ is an upper bound on $f(x)$ for $x < x_0$ and a lower bound on $f(x)$ for $x > x_0$. From the two constraints per function, the constants of $i_l(x)$ and $d_l(x)$ can be established. The result then follows immediately. □

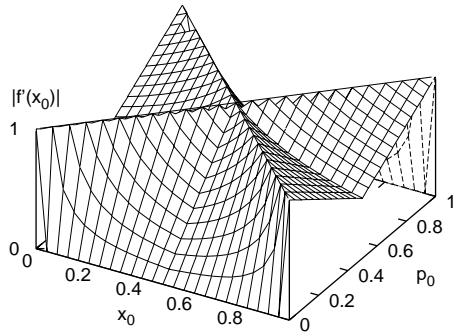

Figure 4: The upper bound on the sensitivity value of a linear sensitivity function, as a function of $x_0$ and $p_0$.

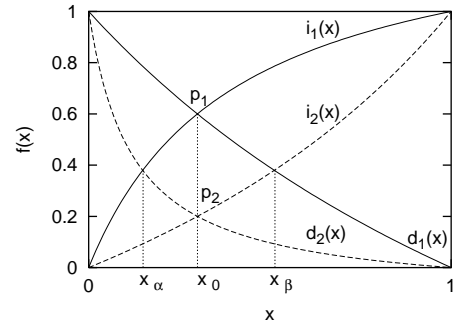

Figure 5: The minimum admissible deviation defined by the space between the hyperbolic bounds for $p_1$ and $p_2$.

An upper bound on the sensitivity value $|f'(x_0)|$ of a linear sensitivity function is readily found to be the maximum of the absolute values of the gradients of the increasing and decreasing linear bounds given in Proposition 3.2. The upper bound on the sensitivity value as a function of $x_0$ and $p_0$ is shown in Figure 4. Note that the sensitivity value of a linear function never exceeds 1; further note that the surface from Figure 4 can be placed underneath that of Figure 2.

We recall once again that, for real-life probabilistic networks, different profiles of evidence are studied. For each such profile, the parameters that will give rise to a linear sensitivity function are readily determined [4]. For example, the parameters of any variable without observed descendants, will show a linear relationship with the probability of interest. Knowledge of where the variable of interest and the observable variables reside in a network, therefore, provides for identifying ranges of profiles for which specific sensitivity functions are linear.

## 4 BOUNDING ADMISSIBLE DEVIATIONS

In the previous section we showed how bounds on sensitivity functions can be exploited to derive bounds on sensitivity values. Often, however, we are interested not in the sensitivity of a probability of interest to parameter variation, but in the effect on the most likely value for the variable of interest. For some parameters, deviation from their original assessment may considerably change the probabilities of a variable's values without inducing a change in the most likely one; for other parameters, variation may have little effect on the probabilities involved and yet change the most likely outcome. The concept of admissible deviation now captures the sensitivity of the most likely value of a variable of interest to parameter variation [6].

We consider a variable of interest $A$. Suppose that $a_1$ is the most likely value of $A$ given the available evidence. We further consider a parameter $x$ with the original value $x_0$ and address the extent to which $x$ can be varied without another value than $a_1$ becoming the most likely value of $A$. From the sensitivity functions for the probabilities of the separate values of $A$, the admissible deviation for $x$ can be computed exactly. In this section, we show that bounds on the admissible deviation can be established without actually knowing these sensitivity functions. For this purpose, we exploit the bounds on the sensitivity functions found in the previous section. In doing so, we again distinguish between hyperbolic and linear sensitivity functions.

### 4.1 HYPERBOLIC FUNCTIONS

We consider a variable of interest $A$ with $n \geq 2$ possible values. Without loss of generality, we assume that $a_1$ is the most likely value of $A$; we then have that $p_1 = \Pr(a_1 \mid e) \geq p_2 = \Pr(a_2 \mid e)$ for some other value $a_2$ of $A$. We further consider a parameter $x$. Now, let $i_k(x)$ and $d_k(x)$ denote the two rectangular hyperbolas from Proposition 3.1 with $p_0$ replaced by $p_k$, $k = 1, 2$, respectively. The sensitivity function that describes the probability $p_1$ of the most likely value $a_1$ of $A$ in terms of the parameter $x$ then is bounded by $i_1(x)$ and $d_1(x)$; the sensitivity function for the probability of $a_2$ is bounded by $i_2(x)$ and $d_2(x)$. Figure 5 serves to illustrate the basic idea. Now, if $p_1 \neq p_2$, there is a space between the bounds through which neither of the sensitivity functions for $p_1$ and $p_2$ can pass. The boundaries of this space define the deviation that is *minimally* allowed for the parameter under study before the value $a_2$ of the variable of interest may become more likely than $a_1$.

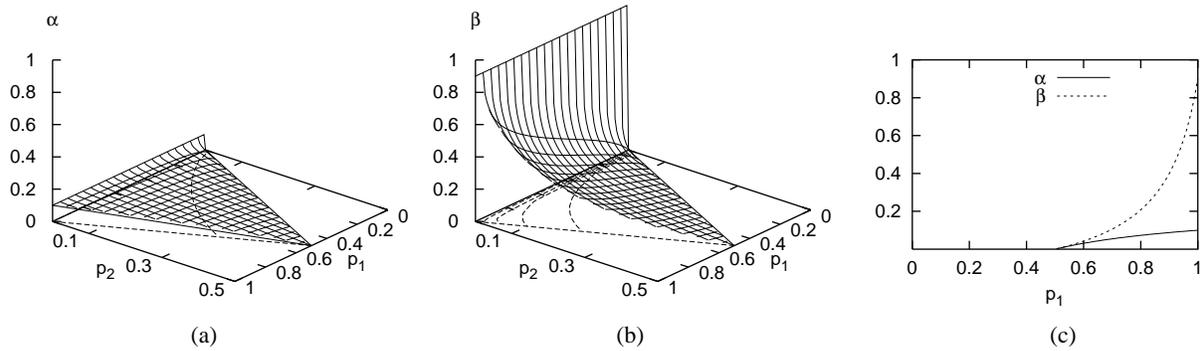

Figure 6: Minimum admissible deviations $(\alpha, \beta)$ for hyperbolic sensitivity functions, plotted as a function of $p_1$ and $p_2$, with $p_1 \geq p_2$ and $p_2 \leq 1 - p_1$, for $x_0 = 0.1$: (a) the minimally admissible shift to smaller values; (b) the minimally admissible shift to larger values; (c) the minimally admissible shifts to smaller values (solid line) and to larger values (dashed line) for a binary variable of interest.

**Proposition 4.1** *Let $x_0$ be the original value of a parameter $x$ and let $p_1$, $p_2$, with $p_1 \geq p_2$, denote the corresponding probabilities of the values $a_1$ and $a_2$ of the variable of interest $A$, respectively. Furthermore, let*

$$Q = \sqrt{p_1 \cdot (1-p_1) \cdot p_2 \cdot (1-p_2)}$$

*Then, the* minimum admissible deviation *from $x_0$ for $a_1$ and $a_2$ equals $(x_0 - x_\alpha, x_\beta - x_0)$, where $x_\alpha = [0, x_0] \cap$*

$$\left\{ -x_0 \cdot \left( \frac{\pm (1-x_0) \cdot Q + (1-p_1) \cdot p_2 \cdot x_0}{(p_1 - p_2) \cdot x_0^2 + (1 - 2 \cdot x_0) \cdot p_1 \cdot (1-p_2)} \right) \right\}$$

*and $x_\beta = [x_0, 1] \cap$*

$$\left\{ -x_0 \cdot \left( \frac{\pm (1-x_0) \cdot Q + (1-p_2) \cdot p_1 \cdot x_0}{(p_2 - p_1) \cdot x_0^2 + (1 - 2 \cdot x_0) \cdot p_2 \cdot (1-p_1)} \right) \right\}$$

**Proof:** We observe that the minimum admissible deviation is computed from the intersections of the bounds on the sensitivity functions for $p_1$ and $p_2$. More in particular, we establish the value $x_\alpha$ from the intersection of $i_1(x)$ and $d_2(x)$. Since hyperbolas have two branches, we find two values for the intersection, one of which lies within the interval $[0, x_0]$. Similar observations hold for $x_\beta$. □

The minimum admissible deviation $(\alpha, \beta)$ from $x_0$ for $a_1$ and $a_2$ indicates that at least a shift by $\alpha$ to smaller values of the parameter and by $\beta$ to larger values, are guaranteed not to change the order of the probabilities $p_1$ and $p_2$ of $a_1$ and $a_2$, respectively. As an example, Figure 6 depicts the minimally admissible shifts to smaller values (a) and to larger values (b) for a parameter with an original value of 0.1, as functions of $p_1$ and $p_2$. Note that $p_1 \geq p_2$ and $p_1 + p_2 \leq 1$; we thus have that $p_2 \leq 0.5$. From the figure, we note that if $p_1 = p_2$ then no deviation from $x_0$ is allowed. The admissible shifts are maximal if $p_2 = 0$. For $x_0 = 0.5$, the minimal shifts allowed would be the same for both smaller and larger values.

We illustrate an example use of the minimum admissible deviation in view of a thorough sensitivity analysis in which various profiles of evidence are studied. We consider Figure 6(c) which shows the intersections of the surfaces from Figures 6(a) and (b) with the plane $p_2 = 1 - p_1$, that is, we consider a binary variable of interest; note that we thus have that $p_1 \geq 0.5$. From the figure we observe for example that, for any profile that results in $p_1 = 0.8$, the minimum admissible deviation from $x_0 = 0.1$ is $(0.075, 0.269)$, indicating that the parameter under study can be varied within the interval $[0.025, 0.369]$ without changing the most likely value of the variable of interest. The same deviation is also admissible for any profile that is known to induce a probability of at least $0.8$. If the plausible interval of variation for the parameter is within the given admissible deviation, then we can safely say that for all profiles that serve to increase the probability of $a_1$, the most likely value of the variable of interest cannot change upon the variation. For profiles that induce a decrease in $p_1$, however, a more detailed analysis is required.

The admissible deviation $(\alpha, \beta)$ for $a_1$ and $a_2$ captures the shifts that can be minimally applied to the original value $x_0$ of the parameter $x$ without changing the order of the probabilities of the values $a_1$ and $a_2$ of the variable of interest $A$. For a binary variable, if $a_1$ is the most likely value given $x_0$, then variation of $x$ within the interval captured by the admissible deviation guarantees that $a_1$ remains to be the most likely value. For non-binary variables, however, this property no longer holds: the value $a_1$ then is guaranteed to remain the most likely value only if the parameter is varied within the intersection of the intervals captured by the admissible deviations for *all* other values $a_2, \ldots, a_n$, $n \geq 2$, of $A$. When a shift beyond the interval defined by the minimum admissible deviation for $a_1$ and $a_2$ is applied to the parameter $x$, then the order of the probabilities of $a_1$ and $a_2$ may change. Note that $a_2$ or in fact another value of $A$, may then become the new most likely value.

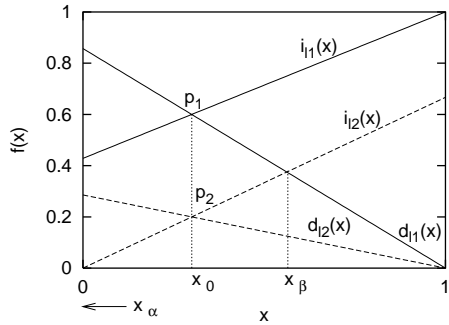

Figure 7: The minimum admissible deviation defined by the space between the linear bounds for $p_1$ and for $p_2$.

## 4.2 LINEAR FUNCTIONS

We consider the same variable of interest $A$ as before, with $a_1$ for its most likely value. We suppose that the probabilities of the values of $A$ relate linearly to the parameter $x$ under study. Now, let $i_{lk}(x)$ and $d_{lk}(x)$ denote the two linear functions from Proposition 3.2 with $p_0$ replaced by $p_k$, $k = 1, 2$, respectively. The sensitivity function that describes the probability $p_1$ of the most likely value $a_1$ of $A$ in terms of the parameter $x$, then is bounded by $i_{l1}(x)$ and $d_{l1}(x)$; the sensitivity function for the probability $p_2$ of the value $a_2$ is bounded by $i_{l2}(x)$ and $d_{l2}(x)$. Figure 7 illustrates the basic idea. If $p_1 \neq p_2$, then the space between the bounds on the sensitivity functions again defines the *minimum* admissible deviation from $x_0$ for $a_1$ and $a_2$.

**Proposition 4.2** *Let $x_0$ be the original value of a parameter $x$ and let $p_1$, $p_2$, with $p_1 \geq p_2$, denote the corresponding probabilities of the values $a_1$ and $a_2$ of the variable of interest $A$, respectively. Then, the* minimum admissible deviation *from $x_0$ for $a_1$ and $a_2$ equals $(x_0 - \max\{x_\alpha, 0\}, \min\{x_\beta, 1\} - x_0)$, where*

$$x_\alpha = \begin{cases} \dfrac{p_2 \cdot x_0}{p_1 - (p_1 - p_2) \cdot x_0} & \text{if } 1 - p_2 > x_0 \geq p_1 \\[1em] \dfrac{x_0}{p_1 - p_2 + 1} & \text{if } x_0 \geq p_1 \text{ and } x_0 \geq 1 - p_2 \\[1em] \dfrac{(1 - p_1) \cdot x_0}{1 - p_2 - (p_1 - p_2) \cdot x_0} & \text{if } 1 - p_2 \leq x_0 < p_1 \\[1em] \dfrac{p_1 - p_2 - x_0}{p_1 - p_2 - 1} & \text{if } x_0 < p_1 \text{ and } x_0 < 1 - p_2 \end{cases}$$

*and $x_\beta$ is defined similarly, with $p_1$ and $p_2$ interchanged.*

**Proof:** We observe that the minimum admissible deviation is computed from the intersections of the bounds on the sensitivity functions for $p_1$ and $p_2$. More specifically, we establish the value $x_\alpha$ from the intersection of $i_{l1}(x)$ and $d_{l2}(x)$, and the value $x_\beta$ from the intersection of $i_{l2}(x)$ and $d_{l1}(x)$. If $x_0 < p_1$ and $x_0 < 1 - p_2$ then $x_\alpha$ may be negative; similarly, if $x_0 \geq p_2$ and $x_0 \geq 1 - p_1$, then $x_\beta$ may be larger than one. Otherwise, $x_\alpha$ and $x_\beta$ are in the $[0, 1]$-interval. The result now follows immediately. □

The minimum admissible deviation again indicates minimally allowed shifts to smaller values and to larger values of the parameter $x$ under study. As an example, Figure 8 depicts the minimally admissible shifts to smaller values (a) and to larger values (b) for a parameter with an original value of 0.8, as functions of $p_1$ and $p_2$. We observe that both surfaces show a level flat. To explain this observation, we recall that the intersection of the linear bounds may fall outside the $[0, 1]$-interval. If the intersection gives $x_\alpha < 0$, for example, then the parameter under study can be varied to zero without inducing a change in the more likely value of $a_1$ and $a_2$. The minimally admissible shift to smaller values then equals $x_0$. For increasing values of $p_1$ and decreasing values of $p_2$, the minimally admissible shift will then remain to be equal to $x_0$, thereby giving rise to the flat of Figure 8(a). If the intersection gives $x_\beta > 1$, then the minimally admissible shift to larger values equals $1 - x_0$.

We now consider Figure 8(c) which shows the intersections of the surfaces from Figures 8(a) and (b) with the plane $p_2 = 1 - p_1$, that is, we consider again a binary variable of interest. We observe that both the minimally admissible shift $\alpha$ to smaller values and the minimally admissible shift $\beta$ to larger values, expressed as functions of $p_1$, show points at which the function is not differentiable. The function that expresses $\beta$ in terms of $p_1$ has such a point at the value of $p_1$ for which $\beta = 1 - x_0$. Note that the value of $\beta$ cannot increase beyond $1 - x_0$ as it already corresponds with a shift to the upper boundary of the $[0, 1]$-interval. The function that expresses $\alpha$ in terms of $p_1$ has a similar point, for which $\alpha = x_0$. The function, moreover, is not differentiable at $p_1 = x_0$. To explain this observation, we note that the minimally admissible shift to smaller values is determined by the intersection of the increasing bound $i_{l1}(x)$ on the linear sensitivity function through $p_1$ and the decreasing bound $d_{l2}(x)$ through $p_2$. We now recall from Proposition 3.2 that the increasing bound $i_{l1}(x)$ is a different function for $x_0 \geq p_1$ and for $x_0 < p_1$; similarly, the decreasing bound $d_{l2}(x)$ differs for $x_0 \geq 1 - p_2 = p_1$ and for $x_0 \leq p_1$. The function that expresses $\alpha$ in terms of $p_1$ therefore is built from three different functions, giving rise to the two points at which the function is not differentiable. In essence, a similar observation holds for the function that expresses $\beta$ in terms of $p_1$. However, since $x_0 = 0.8$ is always larger than $1 - p_1 = p_2$, the increasing bound $i_{l2}(x)$ is described by a single function; also the decreasing bound $d_{l1}(x)$ is captured by a single function. The function expressing $\beta$ therefore is built from two functions, giving rise to just a single point at which it is not differentiable.

We again demonstrate the use of admissible deviations for studying sensitivities. From Figure 8(c), we observe for example that, if $p_1 = 0.8$, then for any evidence profile that induces an increase in this probability, the admissible deviation is at least $(0.3, 0.2)$, indicating that the parameter under study can be varied within the interval $[0.5, 1.0]$ without changing the most likely value of the variable of interest. If the plausible interval of variation for the parameter is within this deviation, then we can safely say that for all profiles that will increase the probability of $a_1$, the most likely value of the variable of interest cannot change upon the variation. For profiles that induce a decrease in the probability $p_1$, however, further analysis is required.

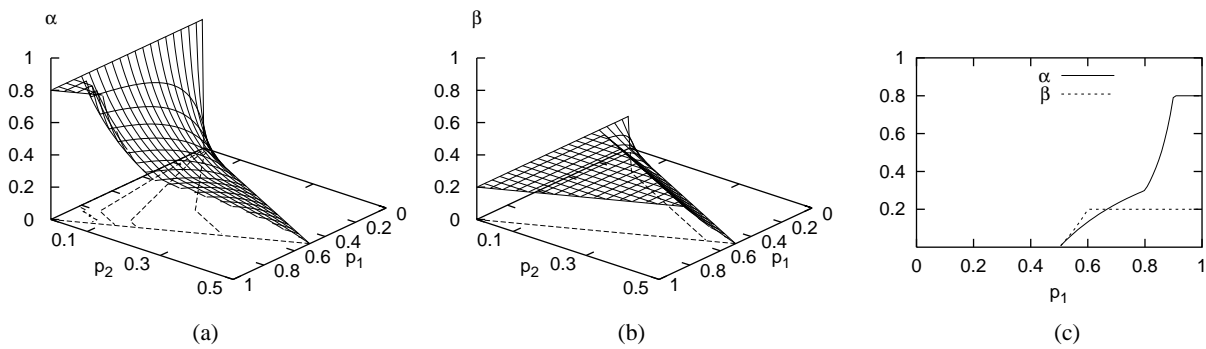

Figure 8: Minimum admissible deviations $(\alpha, \beta)$ for linear sensitivity functions, plotted as a function of $p_1$ and $p_2$, with $p_1 \geq p_2$ and $p_2 \leq 1 - p_1$, for $x_0 = 0.8$: (a) the minimally admissible shift to smaller values; (b) the minimally admissible shift to larger values; (c) the minimally admissible shifts to smaller values (solid line) and to larger values (dashed line) for a binary variable of interest.

## 5  CONCLUDING OBSERVATIONS

Recent results in sensitivity analysis of probabilistic networks showed that bounds can be established on the change in a probability of interest that is occasioned by a shift in a given parameter, without any knowledge of the network at hand. In this paper, we further elaborated on these results and showed that the previously identified bounds are actually built from two hyperbolic functions. We also showed that the bounds can be tightened if the probability of interest is known to relate linearly to the parameter under study. We then exploited these bounds to establish an upper bound on the sensitivity value of a parameter and a lower bound on the deviation that can be applied to the original value of the parameter without changing the most likely value of the variable of interest. We argued that these bounds provide for studying the effects of parameter variation for large ranges of evidence profiles, without the need to perform a complete sensitivity analysis.

To further constrain the bounds on the sensitivity functions for a given probability of interest, we are currently studying the relationship between the four constants in the general form of a sensitivity function and the graphical structure and associated parameters of a probabilistic network. Further insights in the various bounds may also constitute the basis for evidence-invariant bounds on the higher-order sensitivities revealed by real-life probabilistic networks.

## Acknowledgements

This research was (partly) supported by the Netherlands Organisation for Scientific Research (NWO).

## References

[1] E. Castillo, J.M. Gutiérrez, A.S. Hadi (1997). Sensitivity analysis in discrete Bayesian networks. *IEEE Transactions on Systems, Man, and Cybernetics*, vol. 27, pp. 412 – 423.

[2] H. Chan, A. Darwiche (2002). When do numbers really matter? *Journal of Artificial Intelligence Research*, vol. 17, pp. 265 – 287.

[3] H. Chan, A. Darwiche (2002). A distance measure for bounding probabilistic belief change. *Proceedings of the Eighteenth National Conference on Artificial Intelligence*, AAAI Press, Menlo Park, pp. 539 – 545.

[4] V.M.H. Coupé, L.C. van der Gaag (2002). Properties of sensitivity analysis of Bayesian belief networks. *Annals of Mathematics and Artificial Intelligence*, vol. 36, pp. 323 – 356.

[5] U. Kjærulff, L.C. van der Gaag (2000). Making sensitivity analysis computationally efficient. *Proceedings of the Sixteenth Conference on Uncertainty in Artificial Intelligence*, Morgan Kaufmann, San Francisco, pp. 317 – 325.

[6] L.C. van der Gaag, S. Renooij (2001). Analysing sensitivity data from probabilistic networks. *Proceedings of the Seventeenth Conference on Uncertainty in Artificial Intelligence*, Morgan Kaufmann, San Francisco, pp. 530 – 537.